\documentclass[a4paper, 10pt, conference]{ieeeconf}      
\usepackage{FG2026}
\usepackage{xcolor}
\usepackage{amsmath}
\usepackage{amssymb}
\usepackage{graphicx}
\usepackage{booktabs}
\usepackage{tcolorbox}
\usepackage{listings}
\tcbuselibrary{listings, breakable}
\usepackage{multirow}
\usepackage{array}
\usepackage{makecell}
\usepackage{amssymb}
\usepackage{pifont}
\usepackage{stfloats} 
\usepackage{float}

\lstdefinestyle{jsonbox}{
  basicstyle=\ttfamily\scriptsize,
  columns=fullflexible,
  keepspaces=true,
  breaklines=true,
  breakatwhitespace=false,
}

\tcbset{
  promptbox/.style={
    colback=gray!5!white,
    colframe=gray!75!black,
    fonttitle=\bfseries,
    breakable,
    before upper={\setlength{\parskip}{2pt}},
  }
}

\definecolor{jh}{RGB}{0, 102, 204}
\definecolor{hs}{RGB}{204, 0, 0}
\definecolor{hn}{RGB}{0, 255, 0}
\definecolor{jun}{RGB}{0, 0, 200}

\newcommand{\cmark}{\ding{51}}
\newcommand{\xmark}{\ding{55}}

\FGfinalcopy 

\IEEEoverridecommandlockouts                              
\def\FGPaperID{8} 

\title{\LARGE \bf
Ordering Matters: Rank-Aware Selective Fusion for Blended Emotion Recognition
}

\usepackage{fancyhdr}

\begin{document}

\ifFGfinal
\author{\parbox{16cm}{\centering
   {\large Junghyun Lee, Hyunseo Kim, Hanna Jang, and Junhyug Noh$^{*}$}\\
   {\normalsize
   Department of Artificial Intelligence and Software, Ewha Womans University, Seoul, Republic of Korea}}\\
   {\tt\small ejunghyun@ewha.ac.kr, \{khsvv, hanna2129032\}@ewhain.net, junhyug@ewha.ac.kr}\\
   {\small $^{*}$Corresponding author}
}
\else
\author{Junghyun Lee, Hyunseo Kim, Hanna Jang, and Junhyug Noh$^{*}$\\
Ewha Womans University\\
{\tt\small ejunghyun@ewha.ac.kr, \{khsvv, hanna2129032\}@ewhain.net, junhyug@ewha.ac.kr}\\
{\small $^{*}$Corresponding author}
}
\pagestyle{plain}
\fi
\maketitle
\thispagestyle{fancy}

\begin{abstract}
Blended emotion recognition is challenging because emotions are often expressed as mixtures of subtle and overlapping multimodal cues rather than a single dominant signal. We propose a rank-aware multi-encoder framework that selectively combines complementary representations from diverse pre-extracted video and audio encoders. Our method projects heterogeneous encoder features into a shared latent space, estimates sample-wise encoder importance through an attention-based gating module, and fuses only the top-$n$ most informative encoders. To better model blended emotions, we decouple prediction into presence and salience heads and align them through probability-level fusion. We further incorporate feature-level unsupervised domain adaptation without pseudo-labeling to improve robustness under distribution shift. Experiments on the BlEmoRE challenge show that the proposed framework outperforms strong individual encoders and naïve multi-encoder fusion baselines. Our final system ranked 2nd in the competition, supporting the effectiveness of rank-aware selective fusion for fine-grained blended emotion recognition.
\end{abstract}

\section{Introduction}

Emotion recognition has long been studied through theories of basic emotions, which describe affective states such as anger, fear, sadness, disgust, and happiness as distinguishable emotion families \cite{darwin1872expression,ekman1992argument,ekman2011basic}. In real-world settings, however, emotions are often not expressed as a single dominant category. Instead, psychological studies and recent affective computing benchmarks suggest that affective states are frequently \emph{blended}, with multiple emotions co-occurring at different levels of prominence depending on context \cite{oatley1994experience,moeller2018mixed,oh2022specificity,lachmann2026blemore}. This makes blended emotion recognition more challenging than conventional single-label classification, since a model must capture both which emotions are present and how strongly each contributes relative to the others.

This challenge is further amplified by the multimodal nature of emotional expression. Facial behavior, vocal prosody, and other contextual cues provide complementary evidence, but their usefulness is often uneven across samples \cite{lian2023survey,lian2024merbench,khare2024emotion}. As a result, simply using more encoders or aggregating all available features does not necessarily lead to better performance. Different encoders may capture overlapping information, exhibit varying robustness across inputs, or contribute only in specific cases. Prior work on attention-based fusion and conditional computation similarly suggests that not all modalities or experts should be treated equally \cite{priyasad2020attentionfusion,shazeer2017moe}. These observations motivate a selective view of multimodal fusion, in which encoder contributions are ranked and combined adaptively rather than fused uniformly.

In this work, we propose a rank-aware multi-encoder framework for blended emotion recognition. Our method projects heterogeneous video and audio encoder features into a shared latent space, estimates sample-wise encoder importance through an attention-based gating module, and selectively fuses only the top-$n$ most informative encoders. On top of the fused representation, we use dual prediction heads to model emotion presence and salience separately, and align them through probability-level fusion. We also incorporate feature-level unsupervised domain adaptation without pseudo-labeling to improve robustness under distribution shift \cite{ganin2016dann}. Our final system ranked 2nd in the BlEmoRE competition, suggesting that selective fusion and domain-aware training provide an effective framework for fine-grained blended emotion recognition.

Our main contributions are as follows:
\begin{itemize}
    \item We formulate blended emotion recognition as a selective fusion problem, where encoder contributions are ranked dynamically rather than treated uniformly or reduced to a single best backbone.
    \item We propose a rank-aware multi-encoder framework that combines attention-based encoder ranking, top-$n$ selective fusion, and dual-head prediction for emotion presence and salience.
    \item We show that feature-level unsupervised domain adaptation improves robustness under distribution shift and contributes to strong performance on the BlEmoRE challenge.
\end{itemize}

\section{Proposed Method}

\begin{figure*}[t]
    \centering
    \includegraphics[width=\textwidth]{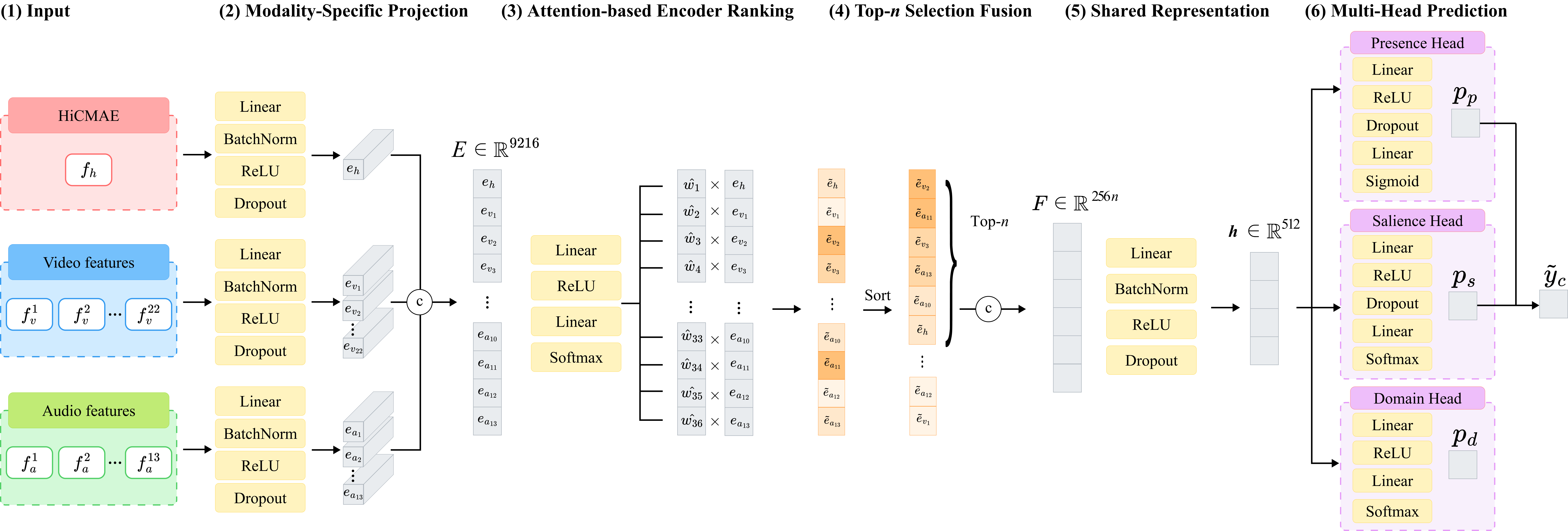}
    \caption{\textbf{Overview of the proposed framework.} Heterogeneous encoder features are first projected into a shared 256-d embedding space. An attention-based gating module estimates sample-wise encoder importance, after which only the top-$n$ encoders are retained for weighted fusion into a 512-d shared representation. Two prediction heads model emotion presence and salience, and their outputs are aligned through probability-level fusion.}
    \label{fig:framework}
\end{figure*}

We address blended emotion recognition using a diverse pool of pre-extracted multimodal encoder features. Let $\mathcal{X}=\{x_1, x_2, \dots, x_M\}$ denote the set of features extracted from $M$ encoders spanning both video and audio streams. Rather than assuming equal encoder contribution or selecting a single best encoder, we estimate sample-wise encoder relevance and selectively fuse only the most informative subset.

\smallskip
\noindent\textbf{Modality-specific projection.}
Because the encoder pool is highly heterogeneous, we first map each encoder feature into a common latent space. For the $i$-th encoder feature $x_i$, we compute
\begin{equation}
e_i = f_i(x_i),
\end{equation}
where $f_i(\cdot)$ is a modality-specific projection block composed of a linear layer, batch normalization, ReLU, and dropout. This yields a 256-dimensional embedding $e_i \in \mathbb{R}^{256}$ for each encoder and reduces dimensional mismatch across encoders while preserving encoder-specific information before fusion.

\smallskip
\noindent\textbf{Attention-based encoder ranking and top-$n$ selective fusion.}
To estimate encoder importance in a sample-adaptive manner, we concatenate the projected embeddings,
\begin{equation}
e = [e_1; e_2; \dots; e_M],
\end{equation}
and feed the resulting vector into a lightweight gating network $g(\cdot)$:
\begin{equation}
w = \mathrm{Softmax}(g(e)),
\end{equation}
where $w=[w_1,\dots,w_M]$ and $\sum_{i=1}^{M} w_i = 1$. The score $w_i$ reflects the relative contribution of encoder $i$ for the current sample.

Given these scores, we retain only the top-$n$ encoders with the largest weights. Let $\mathcal{T}\subset\{1,\dots,M\}$ denote the selected index set. The retained weights are renormalized as
\begin{equation}
\hat{w}_i = \frac{w_i}{\sum_{j \in \mathcal{T}} w_j}, \qquad i \in \mathcal{T},
\end{equation}
and used to compute weighted embeddings
\begin{equation}
\tilde{e}_i = \hat{w}_i e_i, \qquad i \in \mathcal{T}.
\end{equation}
The selected embeddings are then concatenated and mapped into a shared fusion space:
\begin{equation}
h = f_{\mathrm{shared}}\big([\,\tilde{e}_{i_1}; \tilde{e}_{i_2}; \dots; \tilde{e}_{i_n}\,]\big),
\end{equation}
where $f_{\mathrm{shared}}(\cdot)$ denotes a shared fusion layer and $h \in \mathbb{R}^{512}$ is the final fused representation. This selective fusion suppresses less informative signals while emphasizing complementary encoder cues that are most useful for the current sample.

\smallskip
\noindent\textbf{Dual-head prediction for presence and salience.}
Blended emotion recognition requires modeling not only whether an emotion is present, but also how prominent it is relative to other co-occurring emotions. We therefore attach two prediction heads to the shared representation $h$: a \emph{presence} head and a \emph{salience} head. Their logits are
\begin{equation}
z_p = f_p(h), \qquad z_s = f_s(h),
\end{equation}
where $z_p, z_s \in \mathbb{R}^{C}$ and $C$ is the number of emotion classes. The two heads differ only in how their logits are normalized:
\begin{equation}
p_p = \sigma(z_p), \qquad p_s = \mathrm{Softmax}(z_s).
\end{equation}
The presence head captures independent emotion evidence, whereas the salience head emphasizes relative prominence across classes.

Both heads are supervised using the same soft target vector $\mathbf{t}\in[0,1]^C$, where $t_c$ denotes the continuous annotation score for class $c$. We optimize both heads with soft-label cross-entropy:
\begin{equation}
\mathcal{L}_k = -\sum_{c=1}^{C} t_c \log p_k^{(c)}, \qquad k \in \{p,s\},
\end{equation}
and define the task loss as
\begin{equation}
\mathcal{L}_{\mathrm{task}} = \lambda_p \mathcal{L}_p + \lambda_s \mathcal{L}_s,
\end{equation}
where $\lambda_p$ and $\lambda_s$ control the contributions of the two heads. Although binary cross-entropy is a natural alternative for the presence head, we found soft-label cross-entropy to perform better in practice.

\smallskip
\noindent\textbf{Domain-adversarial learning.}
To improve robustness under distribution shift, we optionally incorporate domain-adversarial learning on top of the shared representation. In our setting, the source domain corresponds to the labeled training set, whereas the target domain corresponds to the unlabeled test set containing unseen speakers. A domain classifier $f_d(\cdot)$ is attached to $h$ through a gradient reversal layer \cite{ganin2016dann}. Let $d \in \{0,1\}$ denote the domain label, where $0$ indicates the source domain and $1$ indicates the target domain. The domain logits and probabilities are
\begin{equation}
z_d = f_d(h), \qquad p_d = \mathrm{Softmax}(z_d),
\end{equation}
and the corresponding domain loss is
\begin{equation}
\mathcal{L}_{\mathrm{domain}}
=
-\sum_{k=0}^{1} \mathbf{1}[d=k]\log p_d^{(k)}.
\end{equation}
When enabled, the full training objective becomes
\begin{equation}
\mathcal{L} = \mathcal{L}_{\mathrm{task}} + \lambda_d \mathcal{L}_{\mathrm{domain}},
\end{equation}
where $\lambda_d$ controls the contribution of the domain-adversarial term. Through gradient reversal, the fused representation is encouraged to remain discriminative for emotion prediction while being less sensitive to domain differences.

\smallskip
\noindent\textbf{Probability-level alignment and final decoding.}
During inference, we combine the outputs of the presence and salience heads to obtain the final blended-emotion score:
\begin{equation}
\tilde{y}_c
=
\frac{
p_p^{(c)} \cdot \big(p_s^{(c)}\big)^{\alpha}
}{
\sum_{c'=1}^{C} p_p^{(c')} \cdot \big(p_s^{(c')}\big)^{\alpha} + \epsilon
},
\end{equation}
where $\alpha$ controls the influence of the salience head and $\epsilon$ is a small constant for numerical stability. This formulation allows the presence head to suppress unlikely emotions, while the salience head redistributes probability mass among plausible candidates.

To match the challenge output format, we then apply a simple post-processing step. Emotions with $\tilde{y}_c < \tau_p$ are suppressed; if none survive, only the top-scoring emotion is retained with $100\%$ salience. If \textit{neutral} survives, it is treated as mutually exclusive with blended affective emotions, and only the higher-scoring option is kept. Finally, for the two highest-scoring remaining emotions $e_1$ and $e_2$, we compute
\begin{equation}
r = \frac{\tilde{y}_{e_1}}{\tilde{y}_{e_1} + \tilde{y}_{e_2}},
\end{equation}
and quantize the salience pair as
\begin{equation}
(S_1,S_2)=
\begin{cases}
(70,30), & \text{if } r > 0.61,\\
(30,70), & \text{if } r < 0.39,\\
(50,50), & \text{otherwise.}
\end{cases}
\end{equation}
\section{Experiments}

\begin{table*}[t]
\centering
\caption{Main quantitative results on BlEmoRE.}
\label{tab:main_result}
\renewcommand{\arraystretch}{1.1}
\setlength{\tabcolsep}{4pt}
\resizebox{\textwidth}{!}{%
\begin{tabular}{llccccccc}
\toprule
\multirow{2}{*}{\textbf{Method}} &
\multirow{2}{*}{\textbf{Encoder Setting}} &
\multirow{2}{*}{\textbf{UDA}} &
\multirow{2}{*}{\textbf{Top-$n$}} &
\multicolumn{2}{c}{\textbf{Validation}} &
\multicolumn{3}{c}{\textbf{Test}} \\
\cmidrule(lr){5-6} \cmidrule(lr){7-9}
& & &
& $\mathbf{ACC_{pres}}$ & $\mathbf{ACC_{sal}}$
& $\mathbf{ACC_{pres}}$ & $\mathbf{ACC_{sal}}$ & $\mathbf{ACC_{avg}}$ \\
\midrule
\multirow{5}{*}{Baseline}
& ImageBind                        & -- & -- & $0.290 \pm 0.028$ & $0.130 \pm 0.008$ & 0.261 & 0.087 & 0.174 \\
& ImageBind + WavLM               & -- & -- & $0.345 \pm 0.035$ & $0.170 \pm 0.055$ & 0.327 & 0.114 & 0.221 \\
& HiCMAE                          & -- & -- & $0.298 \pm 0.025$ & $0.180 \pm 0.036$ & 0.268 & 0.180 & 0.224 \\
& Trivial baseline (single emotion)
                                   & -- & -- & $0.077 \pm 0.005$ & $0.000 \pm 0.000$ & 0.074 & 0.000 & 0.037 \\
& Trivial baseline (blend) 
                                   & -- & -- & $0.056 \pm 0.005$ & $0.035 \pm 0.003$ & 0.056 & 0.033 & 0.044 \\
\midrule
\multirow{3}{*}{Ours}
& HiCMAE + 22 video + 13 audio encoders
                                   & \xmark     & \xmark     & $0.402 \pm 0.021$ & $0.221 \pm 0.035$ & 0.428 & 0.168 & 0.298 \\
& HiCMAE + 22 video + 13 audio encoders
                                   & \cmark & \xmark     & $\mathbf{0.442 \pm 0.021}$ & $\mathbf{0.221 \pm 0.035}$ & $\mathbf{0.450}$ & 0.165 & 0.307 \\
& HiCMAE + 22 video + 13 audio encoders
                                   & \cmark & \cmark & $0.434 \pm 0.021$ & $0.212 \pm 0.049$ & 0.423 & $\mathbf{0.201}$ & $\mathbf{0.312}$ \\
\bottomrule
\end{tabular}%
}
\end{table*}

\subsection{Experimental Setup}

We evaluate our method on the BlEmoRE benchmark for multimodal blended emotion recognition~\cite{lachmann2026blemore}. Following the official challenge protocol, the task requires predicting both the \emph{presence} of multiple emotions and their relative \emph{salience} within each sample. We report the official challenge metrics for presence and salience, together with their average.

Our model uses a pool of 36 pre-extracted encoder features, comprising 22 video encoders, 13 audio encoders, and 1 HiCMAE encoder. We compare against strong single-encoder baselines, simple multimodal combinations without structured selection, and trivial challenge baselines. We also evaluate controlled variants with and without top-$n$ selection and feature-level unsupervised domain adaptation (UDA) to assess the contributions of selective fusion and domain-aware training.

For the main paper, we summarize only the key implementation details. Each encoder feature is projected into a 256-dimensional latent space, ranked by an attention-based gating module, and selectively fused into a 512-dimensional shared representation. Two prediction heads model emotion presence and salience, and their outputs are aligned at the probability level. The model is trained with Adam, a learning rate of $3\times10^{-4}$, weight decay of $10^{-3}$, ReduceLROnPlateau scheduling, and early stopping with patience 7. All hyperparameters are selected by cross-validation using the official fold split, after which the final model is retrained on the full training set with the selected configuration. Full architectural and training details are provided in the appendix.

\subsection{Results}

\noindent\textbf{Main results.}
Table~\ref{tab:main_result} reports the main quantitative results. Among the baselines, ImageBind+WavLM provides the strongest overall baseline, while HiCMAE is the strongest single-encoder baseline for salience prediction. In contrast, our rank-aware multi-encoder framework performs better by combining HiCMAE with the full pool of video and audio encoders. Adding UDA improves presence performance while maintaining comparable salience accuracy, and further introducing top-$n$ selective fusion improves test salience and yields the best overall average score. These results suggest that blended emotion recognition benefits from both encoder diversity and structured selection.

\smallskip
\noindent\textbf{Effect of domain-adversarial adaptation.}
Table~\ref{tab:main_result} also shows the contribution of feature-level UDA. Adding UDA to the full encoder pool improves test presence accuracy from $0.428$ to $0.450$ while keeping salience performance comparable ($0.168$ vs.\ $0.165$). The same trend appears in cross-validation, where the mean validation presence score increases from $0.402 \pm 0.021$ to $0.442 \pm 0.021$. Combining UDA with top-$n$ selective fusion yields the best overall average score ($0.312$), indicating that domain adaptation remains beneficial when combined with rank-aware selective fusion.

\begin{table}[t]
\centering
\caption{Ablation of key architectural components.}
\label{tab:ablation_arch}
\resizebox{\linewidth}{!}{
\begin{tabular}{lccc}
\toprule
\textbf{Configuration} & $\mathbf{ACC_{pres}}$ & $\mathbf{ACC_{sal}}$ & $\mathbf{Avg}$ \\
\midrule
Full model                    & $\mathbf{0.434 \pm 0.021}$ & $\mathbf{0.212 \pm 0.049}$ & \textbf{0.323} \\
$-$ Attention    & $0.312 \pm 0.038$ & $0.137 \pm 0.016$ & 0.224 \\
$-$ Dual-head & $0.283 \pm 0.023$ & $0.134 \pm 0.024$ & 0.209 \\
\bottomrule
\end{tabular}
}
\end{table}

\smallskip
\noindent\textbf{Ablation of architectural components.}
Table~\ref{tab:ablation_arch} reports the contribution of key architectural components on 5-fold cross-validation. Replacing attention-based ranking with uniform averaging leads to a large drop in average accuracy ($0.323 \rightarrow 0.224$), highlighting the importance of sample-adaptive encoder selection.
Further removing the dual-head design results in an additional decline ($0.224 \rightarrow 0.209$), indicating that separate modeling of presence and salience also provides complementary benefit.

\begin{figure}[t]
    \centering
    \includegraphics[width=\linewidth]{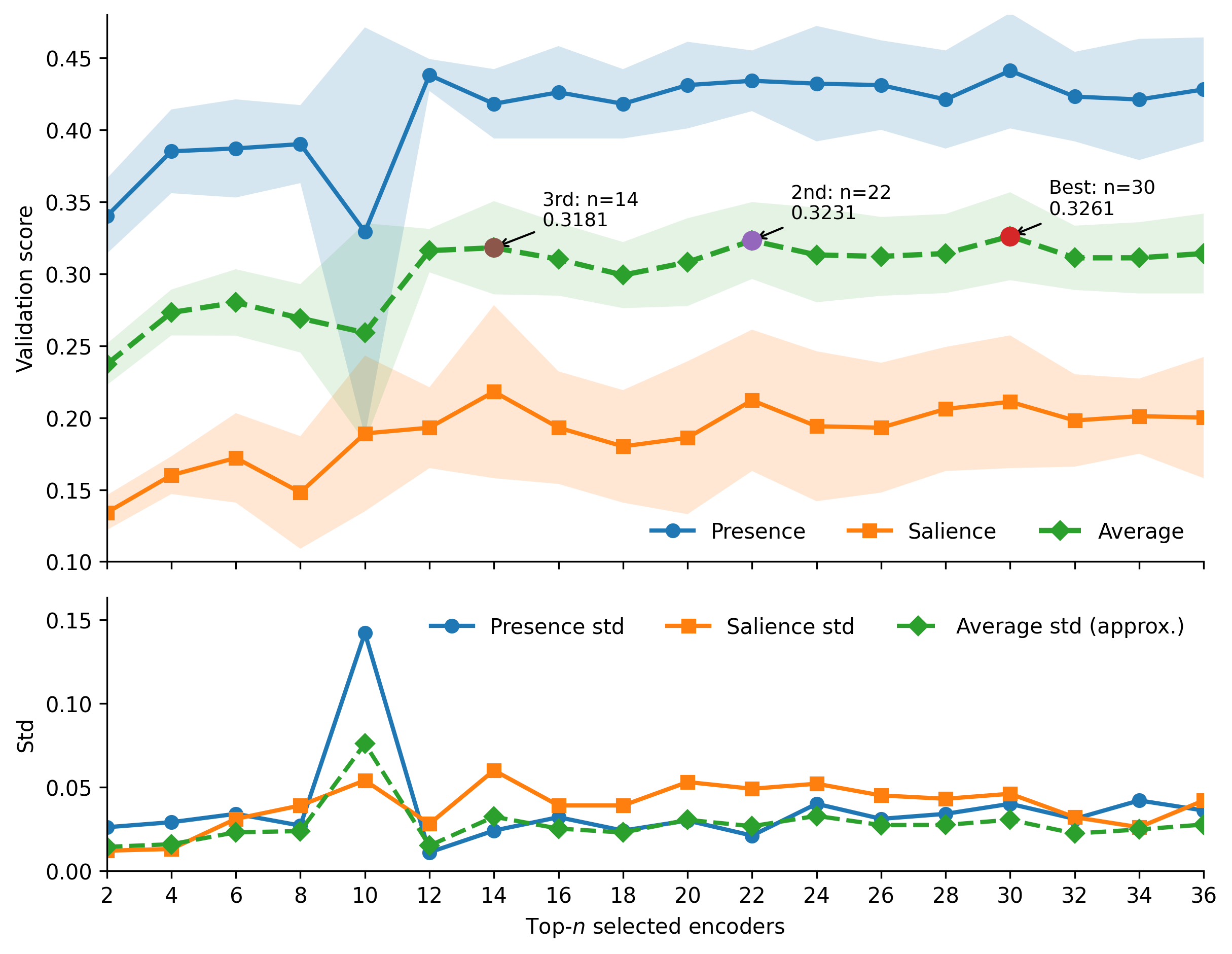}
    \caption{Effect of the number of selected encoders $n$ on validation performance. Intermediate top-$n$ values outperform uniform aggregation over all encoders, supporting the effectiveness of rank-aware selective fusion. Although $n=30$ achieved the highest average score, $n=22$ was selected for the final model because it offered a more stable trade-off across folds with lower variability.}
    \label{fig:topn}
\end{figure}

\smallskip
\noindent\textbf{Effect of top-$n$ selection.}
We next analyze the effect of varying the number of selected encoders. As shown in Figure~\ref{fig:topn}, intermediate top-$n$ values outperform uniform aggregation over all encoder outputs, supporting the effectiveness of rank-aware selective fusion. This suggests that encoder usefulness is not uniform across samples and that retaining only the most informative subset leads to a better fused representation.

Although $n=30$ achieves the highest average validation score, we choose $n=22$ for the final model because it provides a better trade-off between performance and stability across folds, with lower variability. This is consistent with our motivation: the benefit of multi-encoder fusion comes not from indiscriminately increasing the number of encoders, but from selecting a compact subset of complementary signals. Additional cross-validation results for representative top-$n$ values are provided in the appendix.

\begin{figure}[t]
    \centering
    \includegraphics[trim={0 0 0 20pt}, clip, width=\linewidth]{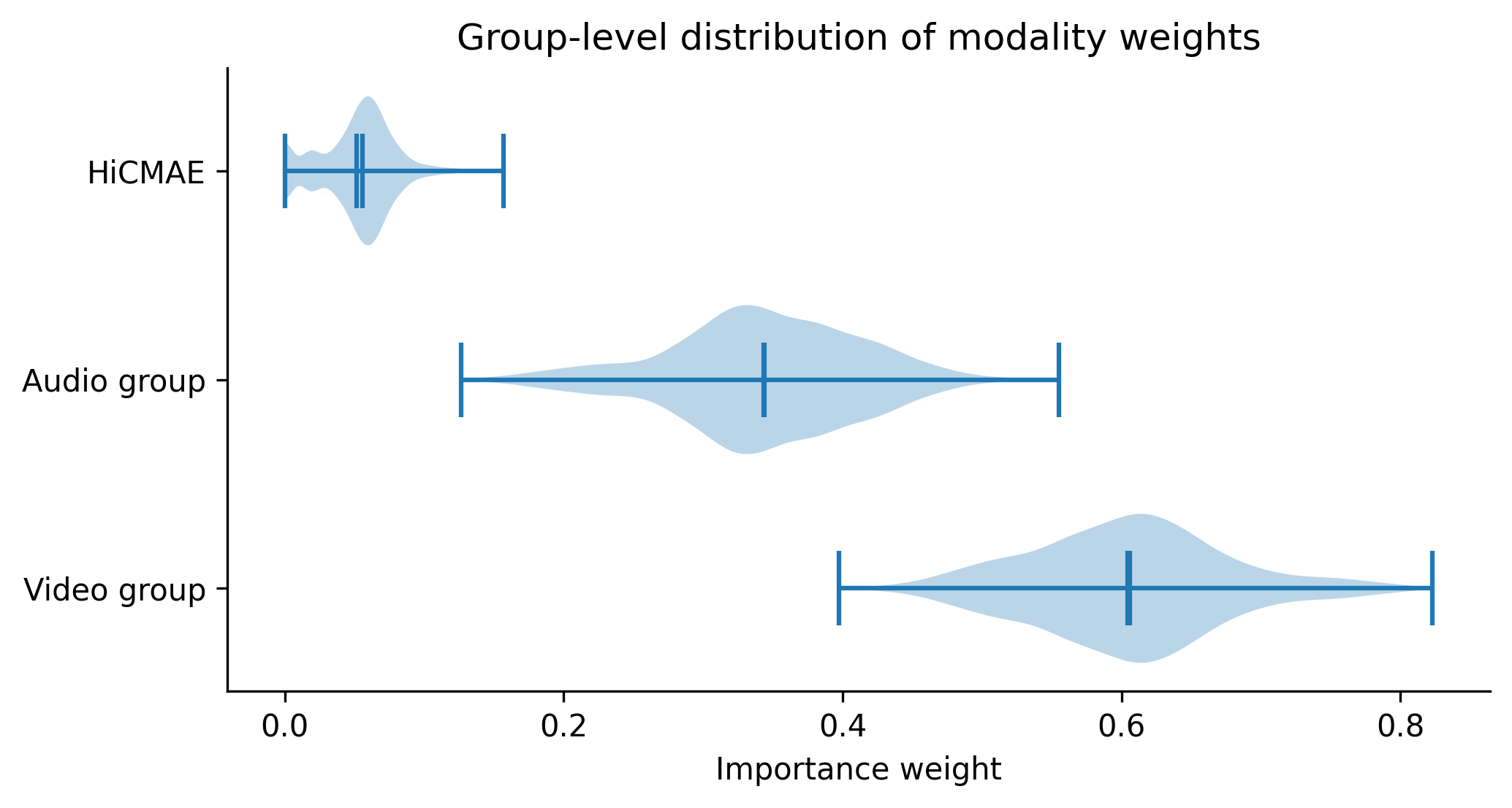}
    \caption{Distribution of modality-group importance scores across samples. The model assigns highly non-uniform importance across encoder groups, with visual encoders dominating the contribution while audio encoders provide complementary signals. This supports the need for rank-aware selective fusion instead of uniform aggregation.}
    \label{fig:violin}
\end{figure}

\smallskip
\noindent\textbf{Analysis of encoder importance.}
We further analyze how encoder contributions are distributed across samples. Figure~\ref{fig:violin} shows that encoder importance is highly non-uniform across modality groups. Visual encoders consistently receive larger weights, while audio encoders provide complementary but typically smaller contributions. This indicates that treating all encoders equally is suboptimal and supports the design of rank-aware selective fusion.

More broadly, the learned weight distribution suggests that encoder usefulness is structured rather than uniform. Some encoder groups contribute consistently across samples, whereas others play a more supplementary role, further supporting sample-adaptive encoder ranking over fixed or uniform fusion strategies.

\section{Conclusion}

We presented a rank-aware multi-encoder framework for blended emotion recognition that estimates sample-wise encoder importance, selectively fuses the top-$n$ encoder features, and jointly models emotion presence and salience. We further incorporated feature-level unsupervised domain adaptation to improve robustness under distribution shift. Experiments on the BlEmoRE challenge show that the proposed framework outperforms strong baselines, and our final system ranked 2nd in the competition. Future work includes adaptive encoder selection, stronger temporal and cross-modal interaction modeling, and tighter integration with large multimodal models.


\newpage

\section*{Acknowledgment}

This work was supported by the Global - Learning \& Academic research institution for Master's·PhD students, and Postdocs (G-LAMP) Program of the National Research Foundation of Korea (NRF) grant funded by the Ministry of Education (No. RS-2025-25442252), and by a research project funded by L4BOX Co., Ltd. The authors also gratefully acknowledge the valuable advice provided by Taewan Kim at Oracle Korea Co., Ltd.



{\small
\bibliographystyle{ieee}
\bibliography{egbib}
}

\clearpage
\appendix


\subsection{Related Work}

\noindent\textbf{Psychological foundations of blended emotions.}
Classic theories describe basic emotions as distinguishable affective families \cite{darwin1872expression,ekman1992argument,ekman2011basic}, whereas constructionist and behavioral studies suggest that emotional experience is often shaped by context and may involve mixed affective states rather than a single isolated category \cite{barrett2007language,lindquist2015role,oatley1994experience,moeller2018mixed,oh2022specificity}. These observations motivate blended emotion recognition, where multiple emotions and their relative prominence must be modeled explicitly \cite{lachmann2026blemore}.

\medskip
\noindent\textbf{Compound, blended, and multimodal emotion recognition.}
Prior work has explored compound and blended emotional expressions in both facial and multimodal settings \cite{du2014compound,zhao2020compound,kollias2023multi,yang2024multimodal,israelsson2023blended}. More broadly, multimodal emotion recognition has shown that robust affect analysis benefits from integrating complementary cues from face, voice, and context \cite{lian2023survey,khare2024emotion,lian2024merbench}. BlEmoRE extends this line of research by explicitly modeling both emotion presence and relative salience in multimodal blended emotion recognition \cite{lachmann2026blemore}.

\medskip
\noindent\textbf{Multimodal backbones, fusion, and selective routing.}
Recent pretraining methods have produced strong backbones for multimodal affect analysis, including VideoMAE-style visual encoders \cite{wang2023videomaev2,sun2023mae}, self-supervised vision encoders such as DINOv2 and DINOv3, and large-scale vision-language and contrastive models such as CLIP, SigLIP, and EVA-based encoders \cite{oquab2023dinov2,simeoni2025dinov3,radford2021learning,zhai2023siglip,sun2023eva}. Facial behavior analysis systems such as OpenFace further provide fine-grained expression cues \cite{baltruvsaitis2016openface,baltrusaitis2018openface,hu2025openface3}.

On the audio side, speech self-supervised models such as HuBERT, WavLM, and wav2vec2 \cite{hsu2021hubert,chen2022wavlm,baevski2020wav2vec}, emotion-oriented encoders such as emotion2vec, and speech foundation models such as Whisper \cite{radford2022whisper} have substantially improved audio representation learning. In addition, audio-text models such as CLAP provide cross-modal representations between sound and language.

Multimodal or affect-oriented encoders such as ImageBind, HiCMAE, and VAEmo further extend cross-modal representation learning \cite{girdhar2023imagebind,sun2024hicmae,cheng2025vaemo}, while recent multimodal large models such as Qwen2.5-VL and InternVL provide strong general multimodal representations \cite{bai2025qwen25vl,zhu2025internvl3,zhao2025humanomni}.

In our framework, these models are used as heterogeneous representation backbones rather than end-to-end predictors. Attention-based fusion has been shown effective for multimodal affect modeling \cite{priyasad2020attentionfusion,mai2024all}, while conditional computation and mixture-of-experts literature suggests that only a subset of experts may be relevant for a given input \cite{shazeer2017moe}. Our method follows this intuition by ranking encoder contributions, selectively retaining the most useful subset, and optionally incorporating domain-adversarial learning \cite{ganin2016dann}.

\subsection{Experimental Setup and Reproducibility Details}

\noindent\textbf{Feature extraction pipeline.}
All encoder features are pre-extracted offline before fusion training and stored as fixed-size representations in \texttt{.npz} format, with one file per encoder per split. For video encoders, the full clip is used without temporal truncation, and each frame is processed independently to obtain frame-level embeddings. For OpenFace-based encoders, facial landmarks, action units, gaze angles, and head pose parameters are extracted from the generated \texttt{.csv} files, retaining only successfully detected frames. For the remaining video encoders, frame-level embeddings are stored as \texttt{.npy} files for each sample.

For audio encoders, the audio stream is first extracted from each video clip as a \texttt{.wav} file, and frame-level embeddings are then obtained using each audio backbone. For HiCMAE, audio-visual features are extracted jointly following the original inference pipeline.

In all cases, frame-level features of shape $(T,D)$ are aggregated into a fixed-size representation of shape $(7D)$ by concatenating seven temporal summary statistics: mean, standard deviation, and the $10$th, $25$th, $50$th, $75$th, and $90$th percentiles. This offline extraction strategy decouples representation learning from fusion training and enables efficient training on pre-computed fixed-size vectors without repeatedly running the encoder backbones.

\medskip
\noindent\textbf{Encoder pool construction.}
Our framework uses a pool of 36 pre-extracted encoder features: 22 video encoders, 13 audio encoders, and 1 HiCMAE encoder. The motivation is to avoid committing to a single backbone and instead exploit encoder diversity across visual, facial, and acoustic representations. The encoder pool includes both general-purpose multimodal or self-supervised backbones and affect-oriented encoders, based on the intuition that different encoders capture complementary emotional cues. In the final system, all encoder features are treated as candidate inputs and are later ranked by the proposed attention-based selection module rather than being manually pruned in advance.

\begin{table}[h]
\centering
\caption{Pre-extracted encoder pool used in the proposed framework.}
\label{tab:encoders}
\renewcommand{\arraystretch}{1.05}
\setlength{\tabcolsep}{8pt}
\resizebox{0.9\linewidth}{!}{
\begin{tabular}{ll}
\toprule
\textbf{Modality} & \textbf{Model} \\
\midrule
\multirow{22}{*}{Video}
 & CLAP \\
 & CLIP \\
 & DINOv2 \\
 & DINOv3 \\
 & EVA Giant CLIP (336) \\
 & EVA Giant CLIP \\
 & EVA Large CLIP \\
 & EVA02 Base \\
 & EVA02 Large \\
 & ImageBind \\
 & OpenFace 2.0 \\
 & OpenFace 3.0 \\
 & SigLIP2 Base (16/384) \\
 & SigLIP2 Giant-Opt (16/384) \\
 & SigLIP2 Large (16/384) \\
 & SigLIP2 SO400M (16/384) \\
 & VideoMAE v2 \\
 & Video Swin Transformer \\
 & InternVL3.5-8B \\
 & InternVL3.5-14B \\
 & InternVL3.5-38B \\
 & InternVL3.5-78B \\
\midrule
\multirow{13}{*}{Audio}
 & WavLM Large \\
 & emotion2vec Base \\
 & emotion2vec+ Base \\
 & emotion2vec+ Large \\
 & emotion2vec+ Seed \\
 & wav2vec2 Base \\
 & wav2vec2 Large (960h) \\
 & wav2vec2 Large Robust (Emotion) \\
 & wav2vec2 Large Robust \\
 & wav2vec2 Large \\
 & wav2vec2 Large XLSR \\
 & HuBERT Large \\
 & Whisper v3 \\
\midrule
Multimodal & HiCMAE \\
\bottomrule
\end{tabular}
}
\end{table}

\medskip
\noindent\textbf{Fusion design.}
Because the encoder backbones produce heterogeneous feature dimensions, each feature is first mapped into a common latent space by a modality-specific projection block. In practice, each encoder feature is projected through a linear layer followed by batch normalization, ReLU, and dropout, producing a 256-dimensional embedding for every encoder stream.

During development, we considered both early-fusion and late-fusion variants, but the final model adopts an attention-based late-fusion design. In this setting, each encoder feature is transformed independently into a 256-dimensional embedding, the embeddings are concatenated, and encoder-wise importance scores are estimated using an attention gate. The model then retains only the top-$n$ encoders through normalized masking, and the selected embeddings are weighted and fused into a shared 512-dimensional representation. This design was preferred because it preserves encoder-specific information before aggregation while enabling sample-wise selective fusion.

\begin{table}[t]
\centering
\caption{Main hyperparameter settings of the proposed framework.}
\label{tab:hyperparams}
\resizebox{0.9\linewidth}{!}{
\begin{tabular}{ll}
\toprule
\textbf{Hyperparameter} & \textbf{Value} \\
\midrule
Learning rate & 3e-4 \\
Weight decay & 1e-3 \\
Optimizer & Adam \\
Scheduler & ReduceLROnPlateau \\
Early stopping patience & 7 \\
Early stopping delta & 0.001 \\
Number of encoder streams & 36 \\
Video encoders & 22 \\
Audio encoders & 13 \\
HiCMAE encoders & 1 \\
Projection dimension & 256 \\
Shared feature dimension & 512 \\
Attention hidden dimension & 128 \\
Top-$n$ selection & 22 \\
Maximum dropout rate & 0.33 \\
Presence loss weight ($w_p$) & 0.68 \\
Salience loss weight ($w_s$) & 0.32 \\
Domain loss weight ($w_d$) & 0.15 \\
Attention temperature & 0.7 \\
Attention temperature min & 0.55 \\
Attention temperature max & 1.25 \\
UDA enabled & Yes / No \\
Gradient reversal weight & 0.3 \\
\bottomrule
\end{tabular}
}
\end{table}

\medskip
\noindent\textbf{Hyperparameter settings and training configuration.}
The main architectural hyperparameters of the final model are summarized in Table~\ref{tab:hyperparams}. Each encoder stream is projected to 256 dimensions, and the fused shared representation has dimension 512. The attention gate is implemented as a two-layer MLP with dimensions $(256 \times M) \rightarrow 128 \rightarrow M$, where $M$ is the number of encoder streams. Both the presence head and the salience head use a 512 $\rightarrow$ 256 $\rightarrow$ $C$ structure, where $C$ is the number of emotion classes.

The model is trained jointly by optimizing the presence and salience heads with weighted losses. Specifically, the implementation computes one loss from the presence logits and another from the salience logits, and combines them using tunable loss weights. During inference, the presence and salience probabilities are combined through probability-level alignment to obtain the final blended-emotion prediction. When UDA is enabled, the shared representation is additionally passed through a gradient reversal layer and a domain classifier, encouraging domain-invariant feature learning under distribution shift. Final hyperparameters are selected based on validation performance.

\subsection{Additional Experimental Results}

\noindent\textbf{Top-$n$ encoder selection.}
Table~\ref{tab:topn_grid} reports cross-validation performance for representative top-$n$ values. Intermediate values consistently outperform both small $n$, which limits encoder diversity, and full aggregation ($n{=}36$), which retains all available signals. Although $n{=}30$ achieves the highest average score, it also shows higher variance in presence accuracy than $n{=}22$, indicating lower stability across folds. We therefore select $n{=}22$ as the final configuration because it provides a better trade-off between average performance and cross-fold stability.

\begin{table}[h]
\centering
\setlength{\tabcolsep}{10pt}
\caption{Effect of top-$n$ encoder selection on 5-fold cross-validation performance.}
\label{tab:topn_grid}
\resizebox{\linewidth}{!}{
\begin{tabular}{cccc}
\toprule
$\boldsymbol{n}$ & $\mathbf{ACC_{pres}}$ & $\mathbf{ACC_{sal}}$ & $\mathbf{Avg}$ \\
\midrule
2  & $0.340 \pm 0.026$ & $0.134 \pm 0.012$ & 0.237 \\
10 & $0.329 \pm 0.142$ & $0.189 \pm 0.054$ & 0.259 \\
16 & $0.426 \pm 0.032$ & $0.193 \pm 0.039$ & 0.310 \\
20 & $0.431 \pm 0.030$ & $0.186 \pm 0.053$ & 0.308 \\
\textbf{22} & $\mathbf{0.434 \pm 0.021}$ & $\mathbf{0.212 \pm 0.049}$ & \textbf{0.323} \\
28 & $0.421 \pm 0.021$ & $0.211 \pm 0.043$ & 0.314 \\
30 & $0.441 \pm 0.040$ & $0.211 \pm 0.046$ & 0.326 \\
36 & $0.428 \pm 0.036$ & $0.200 \pm 0.042$ & 0.314 \\
\bottomrule
\end{tabular}
}
\end{table}

\subsection{Encoder Selection and Representation Analysis}

To better understand the behavior of the proposed model, we analyze encoder selection frequency, fold-wise stability, representative importance distributions, and representational similarity across encoders.

\begin{figure*}[t!]
    \centering
    \includegraphics[trim={0 0 0 20pt}, clip, width=\textwidth]{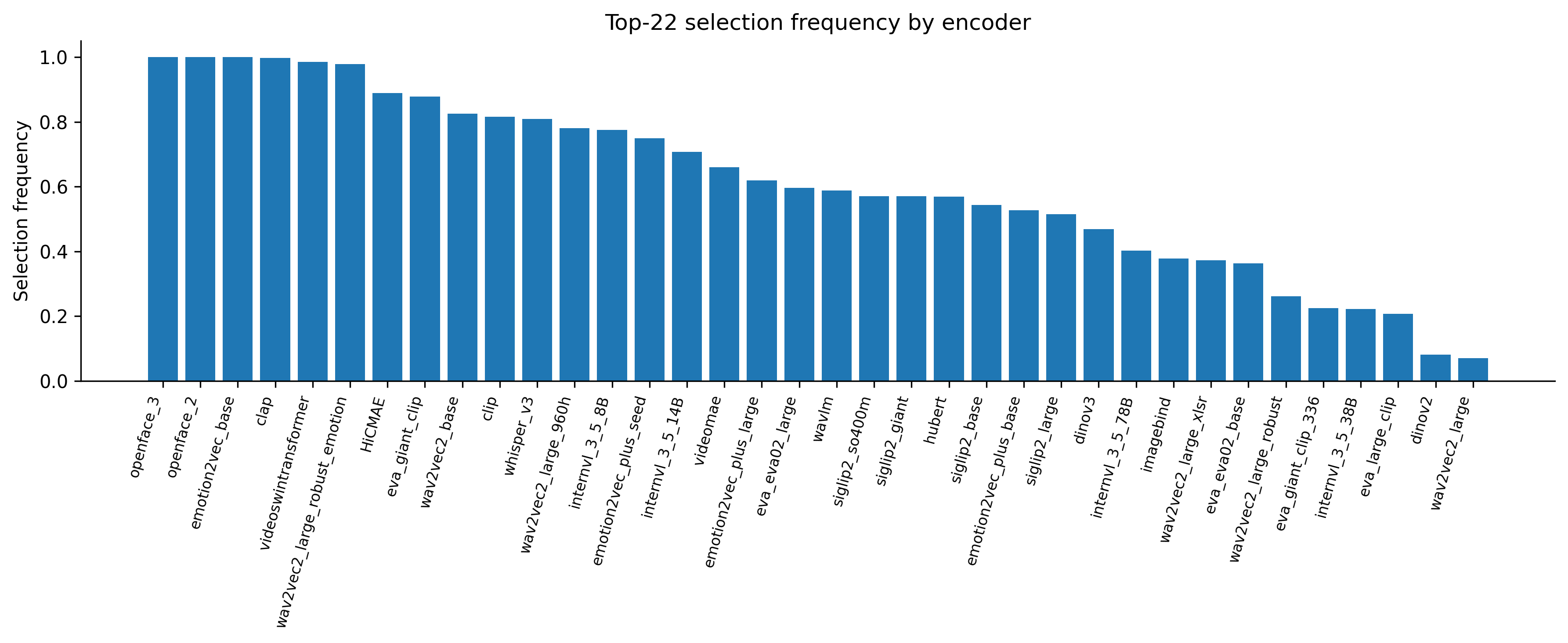}
    \caption{Top-$n$ selection frequency for each encoder. A small subset of encoders is selected in most samples, while many others are used much less frequently. The gradually decaying distribution indicates that encoder usefulness is highly uneven, supporting the need for ranking-based selective fusion.}
    \label{fig:appendix_topk_freq}
\end{figure*}

\medskip
\noindent\textbf{Top-$n$ selection frequency.}
Figure~\ref{fig:appendix_topk_freq} shows how frequently each encoder is selected under the top-$n$ selective fusion strategy. The selection pattern is highly skewed: a small subset of encoders is selected in most samples, whereas many others are used much less frequently. This indicates that encoder utility is not uniform and that the gating module learns a sparse and structured selection policy. The gradually decaying distribution further suggests that encoder usefulness lies on a spectrum rather than a strict useful/non-useful dichotomy.

\begin{figure*}[t!]
    \centering
    \includegraphics[trim={5pt 0 0 20pt}, clip, width=\linewidth]{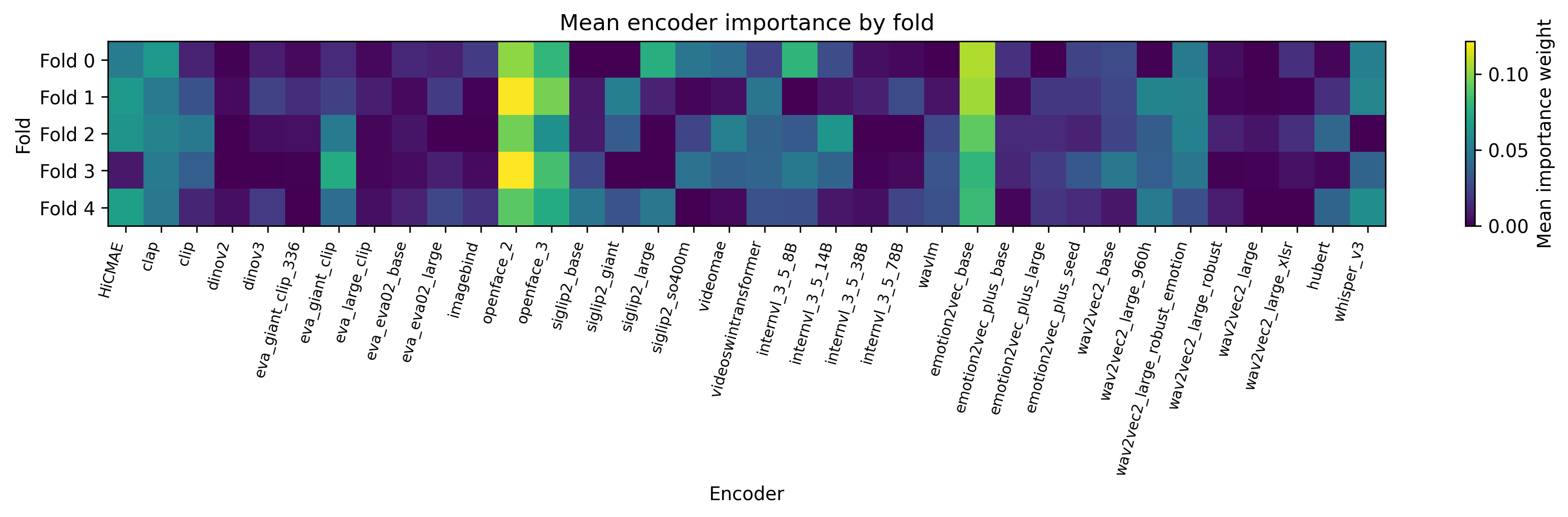}
    \caption{Mean encoder importance across folds. High-importance encoders remain consistently dominant across different folds, while low-importance encoders remain weak, indicating that the learned ranking pattern is stable rather than split-dependent.}
    \label{fig:appendix_fold_heatmap}
\end{figure*}

\medskip
\noindent\textbf{Fold-wise stability of encoder importance.}
Figure~\ref{fig:appendix_fold_heatmap} visualizes the mean encoder importance across folds. The overall ranking pattern remains largely consistent across folds: encoders that receive high importance in one fold also tend to remain highly weighted in others, while low-importance encoders remain weak. Although local variations are still present, the dominant encoder groups remain unchanged. This suggests that the learned ranking is not an artifact of a particular split, but reflects a stable global importance structure.

\begin{figure}[t!]
    \centering
    \includegraphics[trim={0 0 0 20pt}, clip, width=\linewidth]{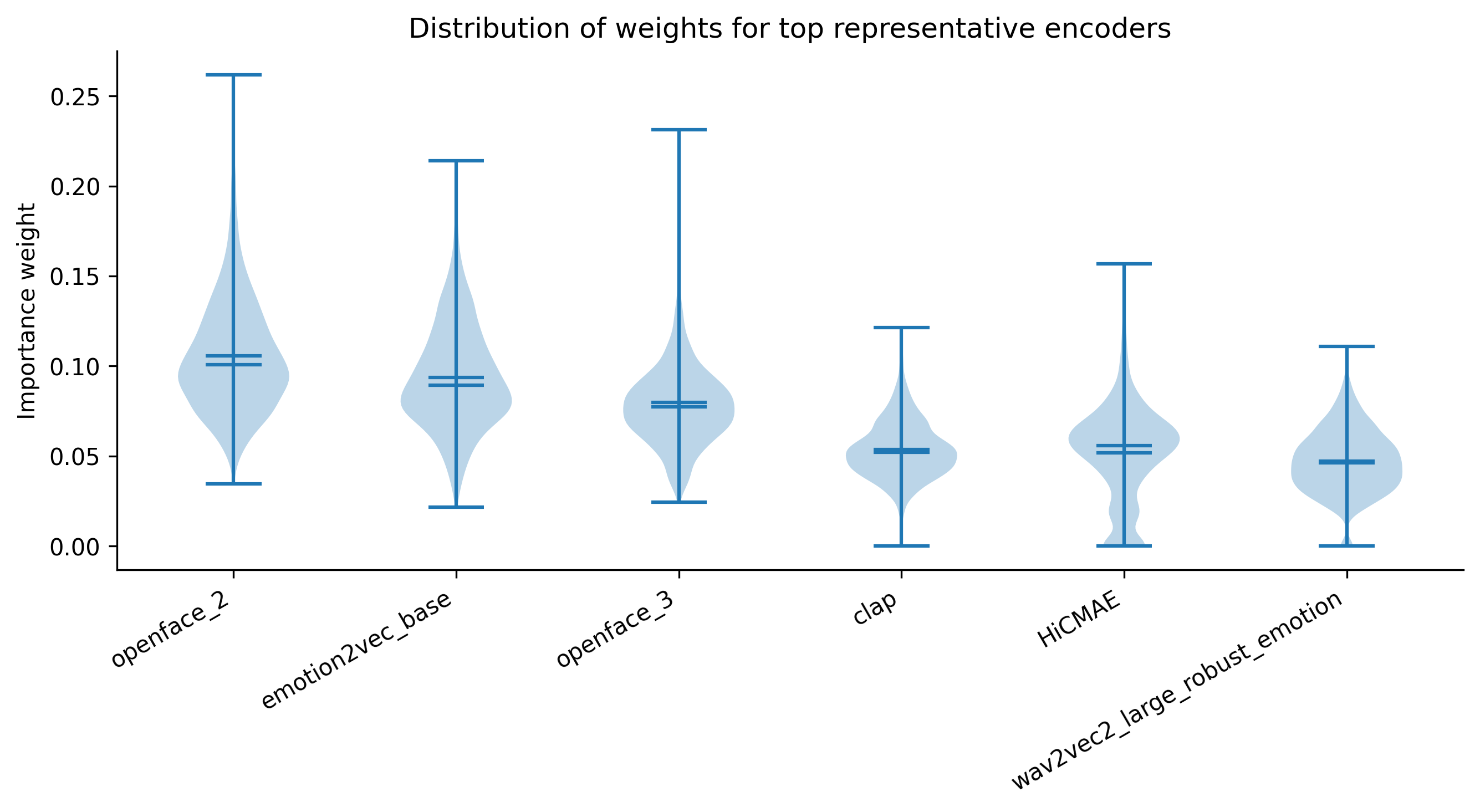}
    \caption{Distribution of importance weights for representative encoders. Facial analysis models (e.g., OpenFace) consistently receive higher weights, while emotion-specific audio encoders contribute moderately and general-purpose encoders receive lower importance. The highly uneven distributions indicate that encoder relevance is strongly skewed, supporting the need for selective fusion.}
    \label{fig:appendix_top_encoder_violin}
\end{figure}

\medskip
\noindent\textbf{Representative encoder distribution.}
Figure~\ref{fig:appendix_top_encoder_violin} shows the distribution of importance weights assigned to a subset of representative encoders across all samples.

The distribution is highly uneven across encoders. Facial analysis models such as OpenFace consistently receive the highest importance, suggesting that visual cues, particularly facial expressions, play a dominant role in blended emotion recognition. Emotion-specific audio encoders such as emotion2vec also receive relatively high importance, but exhibit broader distributions, indicating more sample-dependent contributions. In contrast, general-purpose multimodal encoders such as CLAP and aggregated representations such as HiCMAE receive comparatively lower importance. Overall, these results indicate that only a small subset of encoders consistently dominates the prediction, while others contribute more selectively.

\medskip
\noindent\textbf{Representational similarity analysis.}
To further examine whether the model tends to suppress overlapping encoders, we analyze pairwise similarity between projected encoder embeddings using Linear Centered Kernel Alignment (CKA) \cite{kornblith2019similarity}. CKA is invariant to orthogonal transformations and isotropic scaling, making it suitable for comparing heterogeneous encoder outputs in a shared latent space. For each encoder, we collect 256-dimensional projected embeddings across all training samples and compute pairwise Linear CKA scores for all encoder pairs.

\begin{figure}[t!]
    \centering
    \includegraphics[trim={0 0 40pt 50pt}, clip, width=\linewidth]{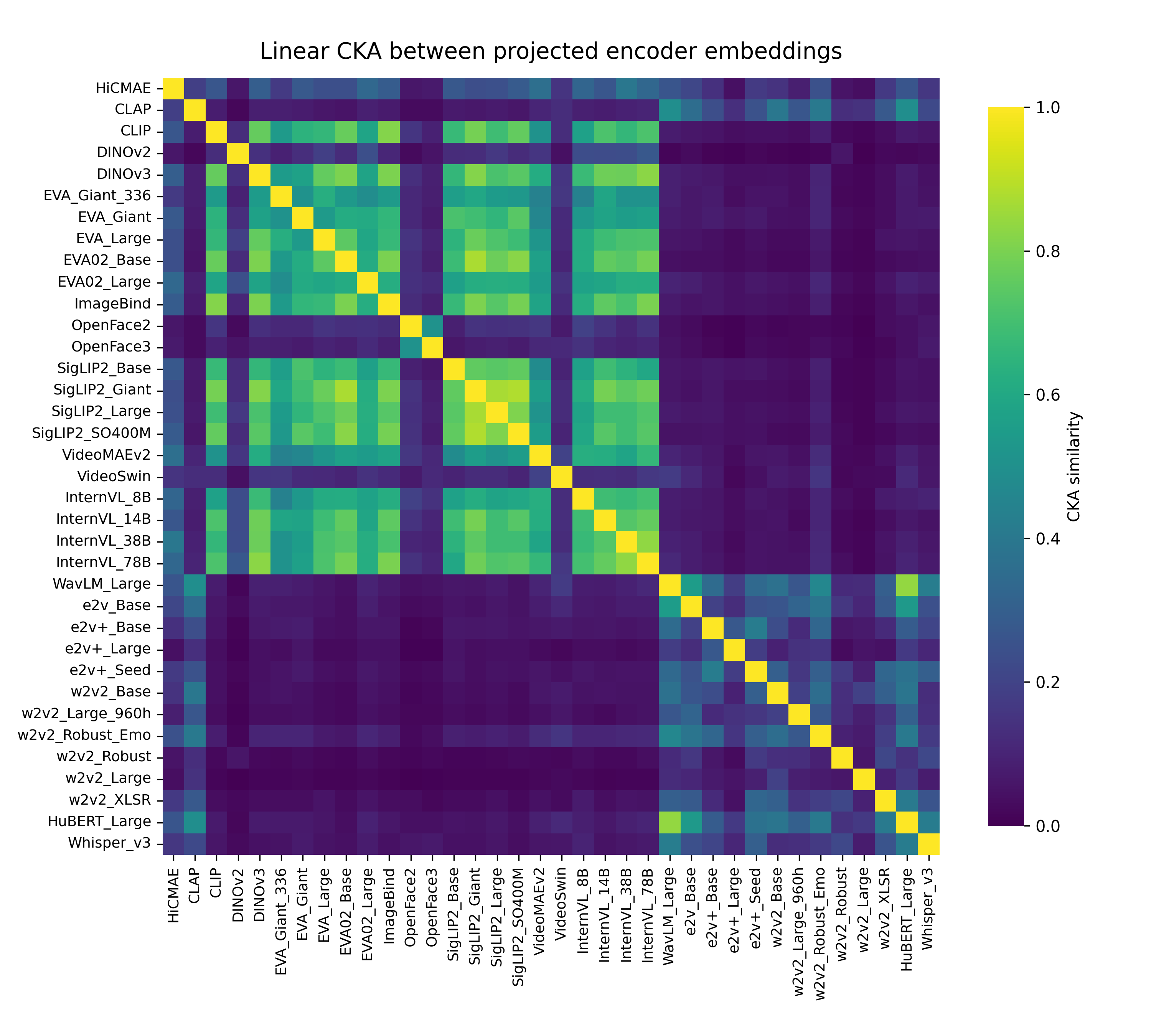}
    \caption{Pairwise Linear CKA similarity between projected encoder embeddings across all training samples. High-similarity clusters are visible within architectural families, while cross-modal video-audio pairs exhibit uniformly low similarity.}
    \label{fig:appendix_cka_heatmap}
\end{figure}

Figure~\ref{fig:appendix_cka_heatmap} shows the resulting similarity matrix. High-similarity clusters are visible within several architectural families, whereas cross-modal video-audio pairs exhibit uniformly low similarity. This pattern suggests that some encoders provide overlapping information, while others contribute more complementary representations.

\medskip
\noindent\textbf{CKA versus co-selection.}
To examine whether the gating module tends to suppress encoders with overlapping representations, we compare pairwise CKA similarity with the co-selection rate, defined as the fraction of samples in which both encoders of a pair are simultaneously retained under top-$n$ selection.

\begin{figure}[t!]
    \centering
    \includegraphics[trim={0 0 0 35pt}, clip, width=0.97\linewidth]{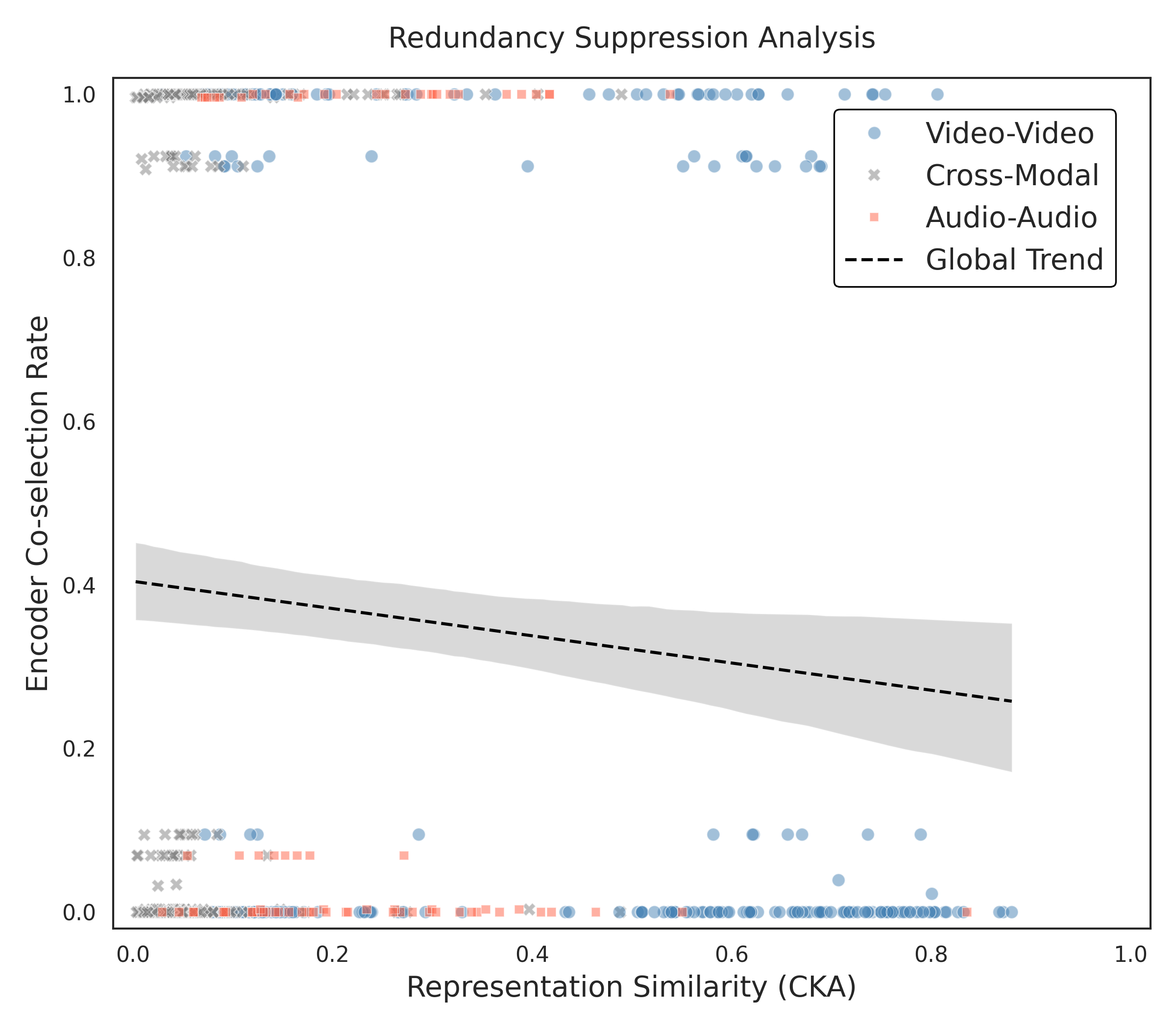}
    \caption{Scatter plot of pairwise CKA similarity versus co-selection rate for all encoder pairs. The negative trend suggests that encoder pairs with higher representational similarity are less likely to be jointly retained under top-$n$ selection.}
    \label{fig:appendix_cka_vs_coselection}
\end{figure}

Figure~\ref{fig:appendix_cka_vs_coselection} shows a negative trend: encoder pairs with higher CKA tend to exhibit lower co-selection rates. This suggests that the gating module tends to down-weight one encoder when a similar counterpart is already retained, rather than selecting highly similar encoders together. Taken together, the CKA and co-selection analyses support the view that rank-aware selective fusion reduces overlap among retained encoder streams.
\end{document}